\documentclass{article}
\usepackage[preprint]{neurips_2025}
\usepackage[utf8]{inputenc}
\usepackage[T1]{fontenc}
\usepackage{microtype}
\usepackage{amsmath,amssymb}
\usepackage{booktabs}
\usepackage{graphicx}
\usepackage{float}
\usepackage{url}
\usepackage{xcolor}
\definecolor{citeblue}{HTML}{176B87}
\definecolor{linkblue}{HTML}{2A5278}
\usepackage{hyperref}
\hypersetup{colorlinks=true,citecolor=citeblue,linkcolor=linkblue,urlcolor=citeblue}
\title{Structured Decoding of World States from Language Model Representations}
\author{Binwen Liu}
\begin{document}
\raggedbottom
\maketitle
\begin{abstract}
Tracking entities through a narrative requires updating changed attributes, holding unchanged ones steady, and composing relations across entities. We study how structured decoders recover these evolving world states from frozen language model representations. Using the bAbI entity-tracking tasks, we construct sentence-level labels for person locations, object holders, and object locations. We compare independent probes, chain conditional random fields (CRFs), and neural sequence encoders against a compositional decoder that predicts person locations and holders before applying a deterministic update rule. On 1,200 test stories, an observation-conditioned CRF with per-sentence MLP potentials and explicit composition reaches $98.1\%$ object-location accuracy, while a BiLSTM-based compositional decoder reaches $99.6\%$. Even under strictly causal forward filtering with approximate soft composition, the MLP decoder attains $97.3\%$. Together, these results show how event-conditioned transitions and explicit relational structure enable accurate, data-efficient trajectory reconstruction, while clarifying how decoder inductive bias and supervision shape probing measurements.
\end{abstract}

\section{Introduction}
Reading a narrative requires maintaining an evolving ledger of facts. When Mary picks up a football and walks into the kitchen, a reader infers that the football is now in the kitchen even though the movement sentence never mentions it; if she later sets the ball down and leaves, their locations decouple again. Because language models process the same sequence of events, recent work has examined how their representations encode entities and changing world states \citep{li2021implicit,kim2023entity,prakash2024finetuning}. Yet answering a single question at the end of a story tests only a final snapshot. Reconstructing a complete step-by-step state trajectory is far more diagnostic: it shows whether decoded states update at the right events, persist unchanged across unrelated sentences, and remain consistent across coupled entities.

Standard probing trains an independent classifier to read a property from a frozen hidden state at a single step \citep{alain2016understanding,belinkov2022probing}. Physical world states, however, follow strong temporal and relational regularities: most variables persist between events, and an object's location depends directly on who carries it. We formulate narrative state reconstruction as sequence labeling on bAbI \citep{weston2016babi}, using a deterministic simulator to assign labels after every sentence for person locations, object holders, and object locations. From sentence-final representations of frozen causal language models, we compare decoders ranging from independent linear and MLP probes to static or observation-conditioned chain CRFs and bidirectional sequence models. Alongside direct object-location predictors, we evaluate a compositional decoder that predicts only the two intermediate state families and recovers object locations through a fixed update rule.

This formulation separates two complementary challenges in state tracking: a temporal decision of when to update a variable versus carry it forward, and a relational step that binds a held object's location to its carrier. Comparing direct and compositional approaches across independent, Markovian, and sequence-level heads shows where each component succeeds or fails, both at single steps and over entire stories. Because the two designs rely on different supervision and structural assumptions, and because structured decoders can themselves aggregate evidence across sentences, we evaluate full-sequence decoding alongside strictly prefix-only filtering, as well as sample efficiency and transfer across story lengths and pretrained models.

\begin{itemize}
  \setlength{\itemsep}{2pt}
  \setlength{\parsep}{0pt}
  \item We construct sentence-level annotations for person locations, object holders, and object locations across bAbI stories, enabling direct evaluation of state updates, silent persistence, and exact trajectory recovery.
  \item We decode person locations and holders separately and combine them with a fixed relational rule. On frozen Qwen2.5-7B features, an observation-conditioned MLP--CRF reaches $98.1\%$ object-location accuracy under this decomposition, compared with $51.5\%$ when trained directly on object locations.
  \item We quantify sensitivity to training data and future context: with 280 training stories the compositional MLP--CRF reaches $81.5\%$ accuracy, and with full training data it reaches $97.3\%$ using prefix-only forward filtering.
\end{itemize}

\begin{figure}[t]
\centering
\includegraphics[width=\linewidth]{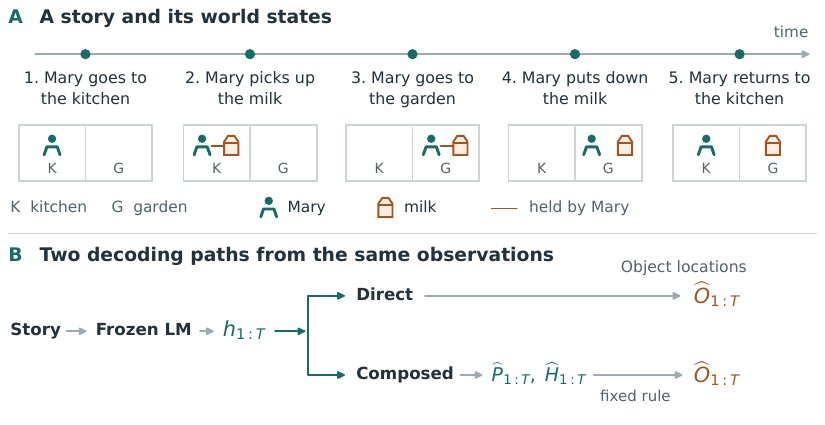}
\caption{World states and decoding paths. (A) Example story showing the simulator's state after each sentence: the milk starts in an unknown location, moves with Mary while held, and stays in the garden after she leaves. (B) Both decoding paths read the same sentence-final hidden states from a frozen causal language model. Direct decoders predict object locations directly, whereas compositional decoders predict person locations and object holders separately and derive object locations via a fixed update rule.}\label{fig:pipeline}
\end{figure}

\section{Related Work}
\textbf{Entity tracking and language model representations.} The bAbI benchmark was designed to test reasoning over chains of supporting facts \citep{weston2016babi}. Subsequent work has studied whether language models form implicit representations of world state \citep{li2021implicit}, how they track entity attributes across context \citep{kim2023entity}, and which internal mechanisms change during fine-tuning \citep{prakash2024finetuning}. Rather than evaluating answers only at selected question points, we reconstruct complete trajectories over every sentence in a story and compare direct prediction against decomposition through intermediate variables. Tracking each decoder's receptive field (distinguishing per-sentence classifiers from chain heads and bidirectional sequence encoders) also separates evidence readable from a single hidden state from information integrated across multiple sentences.

\textbf{Probing and structured prediction.} Linear and shallow probes provide a standard lens onto frozen representations \citep{alain2016understanding}, though interpreting probe performance requires care regarding decoder capacity and task structure \citep{hewitt2019designing,belinkov2022probing}. To make temporal and relational computation explicit rather than leaving it implicit inside a complex probe, we pair frozen representations with chain CRFs \citep{lafferty2001conditional} and a deterministic relational factor, alongside neural sequence models \citep{hochreiter1997long,vaswani2017attention,lample2016neural}. This probabilistic formulation connects directly to hidden Markov models and factor-graph message passing \citep{rabiner1989tutorial,kschischang2001factor}, letting us compare exact MAP paths, smoothed marginals, and causal forward filtering without altering the language model.

\section{Methods}
\subsection{States and observations}
Consider a story of $T$ event sentences. After sentence $t$, let $P_{p,t}$ denote person $p$'s location, $H_{o,t}$ the holder of object $o$, and $O_{o,t}$ the object's location. Each story involves four people, three objects, and six rooms; location variables include an unknown state, and holder variables include nobody. A deterministic simulator assigns labels immediately after every sentence, including steps where an entity is not mentioned. A frozen causal language model reads the story following a fixed introductory prompt, and $h_t$ denotes its hidden state at the final punctuation token of sentence $t$. Because the language model is causal, $h_t$ depends only on the story prefix through $t$.

An example clarifies how the three state families interact. Suppose Mary is in the kitchen, picks up the football, and then travels to the garden. After the pickup, $H_{\mathrm{football},t}=\mathrm{Mary}$ and the football takes her current location (the kitchen). When she next moves to the garden, $P_{\mathrm{Mary},t}$ updates to garden and the held ball moves with her, even though the movement sentence does not mention it. If she subsequently puts the football down, its holder reverts to nobody while its location stays in the garden as Mary moves elsewhere. From the collected representations $h_{1:T}$, a direct decoder predicts $O_{o,1:T}$ directly, whereas a compositional decoder predicts $P_{p,1:T}$ and $H_{o,1:T}$ separately before applying a fixed update rule to obtain object locations.

\begin{figure}[t]
\centering
\includegraphics[width=\linewidth]{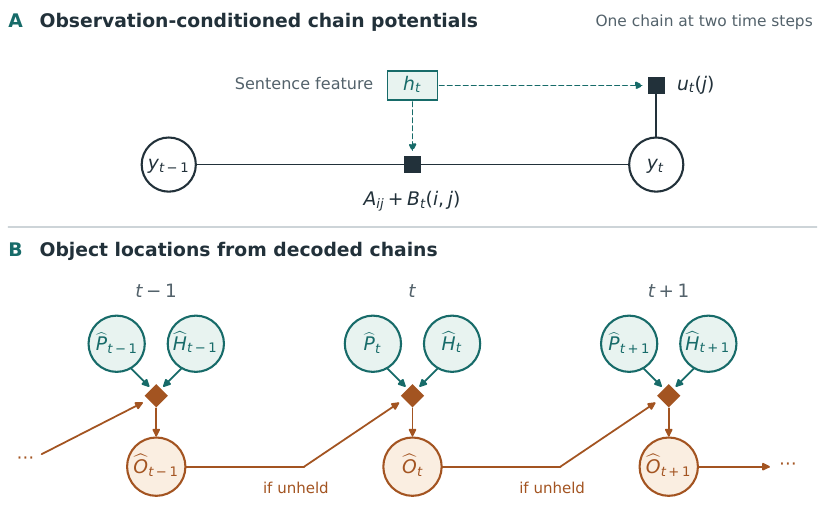}
\caption{Temporal and relational structure of the decoder. (A) Factor graph of the temporal chain decoders, where circles denote state variables, squares denote CRF factors, and dashed arrows indicate observation conditioning in unary and pairwise potentials; separate models score person-location and holder chains. (B) Relational composition (diamonds) combines decoded person locations with an object's holder, carrying the previous object location forward when unheld. Soft composition applies the same update to chain marginals.}\label{fig:decoder}
\end{figure}

\subsection{Chain decoders}
For a single entity's state sequence $y_{1:T}$, a chain CRF assigns the score
\begin{equation}
 s(y,h)=\pi_{y_1}+\sum_{t=1}^{T}u_t(y_t)+\sum_{t=2}^{T}\left[A_{y_{t-1},y_t}+B_t(y_{t-1},y_t)\right],
 \label{eq:crf}
\end{equation}
yielding the conditional probability $p(y\mid h)=\exp s(y,h)/Z(h)$. An encoder maps the observation sequence $h_{1:T}$ to unary scores $u_t$ and, in the observation-conditioned variant, pairwise scores $B_t$. Within each target family, entity slots use separate output projections while sharing static transition parameters.

We evaluate four encoder architectures (linear, per-sentence MLP, bidirectional LSTM, and bidirectional Transformer) across three output structures. An independent classifier omits the chain terms altogether, normalizing $u_t$ over states at each step, training with cross-entropy, and selecting the highest-scoring state per position. A static CRF retains the learned transition matrix $A$ with $B_t=0$, capturing an overall tendency for states to persist across adjacent sentences. Narrative states, however, switch abruptly on action sentences and stay unchanged everywhere else, and a constant matrix $A$ cannot tell a movement sentence apart from an unrelated one. The conditional-transition CRF adds $B_t(i,j)$ computed from the encoder output at step $t$, allowing transition scores to adapt to each event while preserving the shared prior $A$.

These models also differ in what story context informs step $t$. With a per-sentence MLP encoder, both $u_t$ and $B_t$ depend solely on $h_t$. With a BiLSTM or Transformer encoder, scores at step $t$ come from a bidirectional pass over $h_{1:T}$ and can therefore draw on future sentences. Separating encoder context from chain inference identifies which variants operate strictly on a story prefix.

\paragraph{Post-hoc temporal baseline.} We also pair an independently trained classifier with an HMM whose initial and transition probabilities are estimated from training labels using add-one smoothing. Given the classifier's predicted probability $q_t(j)$ and the estimated class prior $\rho(j)$, we use $\log q_t(j)-\log\rho(j)$ as emission scores for Viterbi decoding, separating learned local classification from an empirical transition model.

\subsection{Compositional reconstruction}
As illustrated in Figures~\ref{fig:pipeline} and~\ref{fig:decoder}, the compositional decoder trains one model for the four person-location chains and another for the three holder chains. Given discrete decoded paths $\widehat P_{p,t}$ and $\widehat H_{o,t}$, hard composition updates object locations recursively:
\begin{equation}
 \widehat O_{o,t}=\begin{cases}
 \widehat P_{\widehat H_{o,t},t} & \widehat H_{o,t}\ne\text{nobody},\\
 \widehat O_{o,t-1} & \widehat H_{o,t}=\text{nobody},
 \end{cases}\qquad \widehat O_{o,0}=\text{unknown}.
 \label{eq:compose}
\end{equation}
To retain uncertainty from the intermediate chains, soft composition instead tracks an approximate object-location distribution $q_{o,t}(\ell)$ from the chain marginals:
\begin{equation}
 q_{o,t}(\ell)=p(H_{o,t}=\text{nobody})q_{o,t-1}(\ell)+\sum_p p(H_{o,t}=p)p(P_{p,t}=\ell).
 \label{eq:soft}
\end{equation}
The first term carries the object's previous location distribution forward when it is unheld, while the second sums over potential holders and their predicted locations. Because Equation~\eqref{eq:soft} treats the chain marginals as independent messages, it is a factorized approximation rather than an exact joint posterior over all variables, and the same recurrence applies directly to forward-filtered probabilities. Compositional models train solely on person-location and holder labels (using object-location labels only at evaluation), whereas direct models train on object-location labels directly.

\subsection{Learning and inference}
Each CRF is trained by minimizing the negative log probability of the gold sequence. We compute the partition function $Z(h)$ via the forward recursion, letting $\alpha_t(j)$ denote the log-sum of scores of partial paths ending in state $j$, with $\alpha_1(j)=\pi_j+u_1(j)$ and, for $t>1$,
\begin{equation}
 \alpha_t(j)=u_t(j)+\operatorname{LSE}_{i}\!\left[\alpha_{t-1}(i)+A_{ij}+B_t(i,j)\right],
 \label{eq:forward}
\end{equation}
where $\operatorname{LSE}$ is log-sum-exp and $\log Z(h)=\operatorname{LSE}_{j}\alpha_T(j)$. For $C$ possible states, dynamic programming takes $O(TC^2)$ operations per chain, with padded story steps excluded by a mask, and gradients are computed via automatic differentiation.

At test time, Viterbi decoding yields the single highest-scoring path per chain. We also evaluate forward--backward inference for smoothed marginals over the whole sequence, as well as forward filtering, which uses only potentials up to step $t$. Paired with a causal per-sentence potential encoder, forward filtering followed by Equation~\eqref{eq:soft} produces an object-location distribution after each sentence using only the story prefix.

\section{Experiments}
\paragraph{Data.} We use the English 10k versions of bAbI tasks 2 and 3. Pooling the original training and test splits within each task and deduplicating by full event sequence, we randomly repartition at the story level into 2,800 training (135,872 event sentences), 400 validation (19,798 sentences), and 1,200 test stories (56,216 sentences; 168,648 object-location labels). Task 2 stories average roughly 22 sentences (maximum 88), whereas task 3 stories average roughly 74 sentences (maximum 320). As a sanity check on our simulator, its labels match all 11,000 original current-object-location question answers in task 2 (task 3's historical questions are outside this check), and feeding gold intermediate states into Equation~\eqref{eq:compose} reproduces every test object-location label exactly.

\paragraph{Features and models.} Unless otherwise noted, hidden representations come from layer 14 of frozen Qwen2.5-7B-Instruct \citep{qwen2025qwen}, standardized using training-set statistics. To test generality across language models, we also extract middle-layer features from Qwen2.5-0.5B, 1.5B, and 3B, as well as Llama-3.1-8B-Instruct \citep{grattafiori2024llama}. Across both direct and compositional setups, we evaluate linear probes, a per-sentence MLP (single 512-unit hidden layer), a BiLSTM (256 hidden units per direction), a two-layer Transformer (4 heads, width 256), and the post-hoc HMM baseline built from training-label transition counts \citep{rabiner1989tutorial}.

\paragraph{Training and model selection.} Decoders are trained with AdamW using a batch size of 32 stories for up to 60 epochs, with validation-based early stopping (patience 8). Main results report means and standard deviations across three random seeds; for extended direct BiLSTM and Transformer baselines, we increase the budget to 300 epochs (patience 25). Each decoder is selected by step accuracy on the validation split. In the compositional pipeline, the person-location and holder models are trained and selected independently on their respective state labels, leaving object-location labels untouched until evaluation. Reported parameter counts exclude frozen language-model weights.

\paragraph{Evaluation metrics.} Our primary metric is step-level accuracy across all object--sentence pairs, including unknown states. To measure trajectory-level consistency, we also report whole-story exact match, which requires every object's location to be predicted correctly at every valid step in a story. To disentangle state updates from inertia, we compute accuracy separately on change steps (where the gold state differs from the preceding step) and persistence steps (where it remains unchanged), excluding the first sentence since it lacks a prior state, along with the number of predicted state changes per 100 transitions. Main CRF numbers reflect Viterbi decoding; note that bidirectional sequence encoders and full-sequence CRF inference can access future sentences.

\paragraph{Comparison protocol.} All approaches share identical frozen representations, story partitions, and evaluation labels. Direct decoders train on object-location labels, whereas compositional decoders train on person-location and holder labels and combine them via Equation~\eqref{eq:compose}; consequently, our main comparison contrasts complete decoding pipelines (differing in supervision, parameter count, and domain rule) rather than a single isolated factor. In data-efficiency experiments, the random seed controls both training-set subsampling and decoder initialization, while validation and test splits remain fixed.

\section{Results}
\subsection{Structured reconstruction of object locations}
\begin{figure}[t]
\centering
\includegraphics[width=\linewidth]{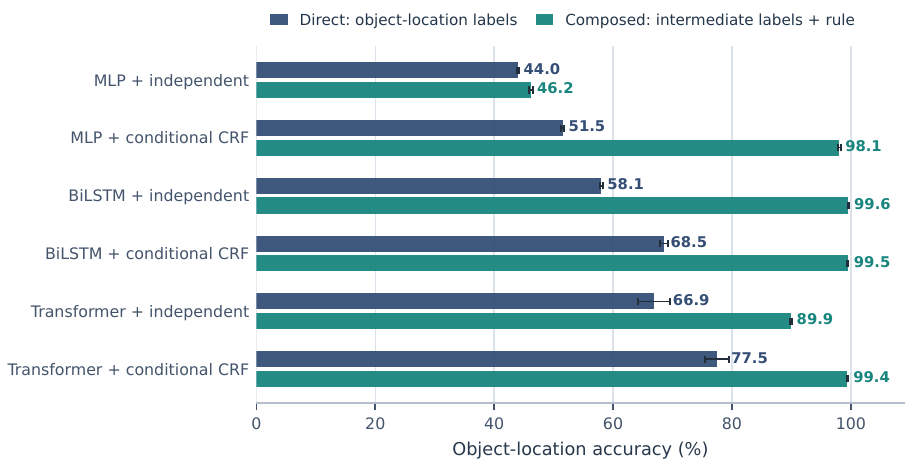}
\caption{Object-location reconstruction from frozen Qwen2.5-7B layer-14 representations. Bars and error bars show means and standard deviations across three seeds ($n=3$). Compositional systems predict person locations and holders before applying a fixed composition rule; direct systems predict object locations directly.}\label{fig:main_results}
\end{figure}
Figure~\ref{fig:main_results} and Table~\ref{tab:main} compare direct and compositional decoders on the 1,200 test stories using Qwen2.5-7B layer-14 features. Predicting object locations directly proves difficult across architectures. A linear probe reaches only $42.0\%$ step accuracy with $0.0\%$ whole-story exact match, and upgrading to an MLP with conditional CRF transitions raises step accuracy to $51.5\%$ ($1.7\%$ story exact match). Even sequence encoders struggle under direct supervision: a BiLSTM with independent outputs scores $58.1\%$, and a two-layer Transformer with conditional transitions reaches $77.5\%$ step accuracy and $24.2\%$ story exact match.

Factoring the problem through person locations and object holders changes this picture sharply, provided the decoder models transitions across sentences. On its own, a compositional MLP with independent per-step heads scores only $46.2\%$ ($0.1\%$ story exact match). Adding a static transition matrix $A$ raises accuracy to $67.4\%$ ($3.5\%$ story), reflecting the benefit of encoding persistence, but still fails when events alter states. Introducing observation-conditioned transitions ($B_t$) bridges that gap: the compositional MLP--CRF jumps to $98.1\%$ step accuracy and $82.8\%$ whole-story exact match. Sequence encoders perform strongly even without a transition head: a compositional BiLSTM with independent outputs reaches $99.6\%$ step accuracy ($92.3\%$ story exact match), while adding conditional CRFs brings BiLSTM and Transformer whole-story exact match to $93.3\%$ ($99.5\%$ step accuracy) and $92.3\%$ ($99.4\%$ step accuracy), respectively. As noted in our comparison protocol, these gains reflect complete compositional pipelines with intermediate supervision and a fixed update rule rather than a single architectural change in isolation.

\begin{table}[t]
\centering\small
\caption{Object-location reconstruction from Qwen2.5-7B layer-14 features (mean $\pm$ standard deviation across three seeds; whole-story exact match reports means). Compositional methods use intermediate-state supervision and a fixed update rule.}\label{tab:main}
\begin{tabular}{llrrr}
\toprule
Approach & Decoder & Accuracy (\%) & Story (\%) & Params.\\
\midrule
Direct & Linear + independent & $42.0\pm0.1$ & 0.0 & 0.08M\\
Direct & MLP + conditional CRF & $51.5\pm0.3$ & 1.7 & 1.92M\\
Direct & BiLSTM + independent & $58.1\pm0.2$ & 1.0 & 1.98M\\
Direct & Transformer + independent, extended & $72.0\pm1.8$ & 7.6 & 1.98M\\
Direct & Transformer + conditional CRF & $77.5\pm2.0$ & 24.2 & 2.02M\\
\midrule
Composed & MLP + independent & $46.2\pm0.3$ & 0.1 & 3.69M\\
Composed & MLP + static CRF & $67.4\pm0.3$ & 3.5 & 3.69M\\
Composed & MLP + conditional CRF & $98.1\pm0.3$ & 82.8 & 3.83M\\
Composed & BiLSTM + independent & $99.6\pm0.1$ & 92.3 & 3.96M\\
Composed & BiLSTM + conditional CRF & $99.5\pm0.1$ & 93.3 & 4.10M\\
Composed & Transformer + conditional CRF & $99.4\pm0.1$ & 92.3 & 4.03M\\
\bottomrule
\end{tabular}
\end{table}

\subsection{Intermediate states and inference}
\begin{figure}[t]
\centering
\includegraphics[width=\linewidth]{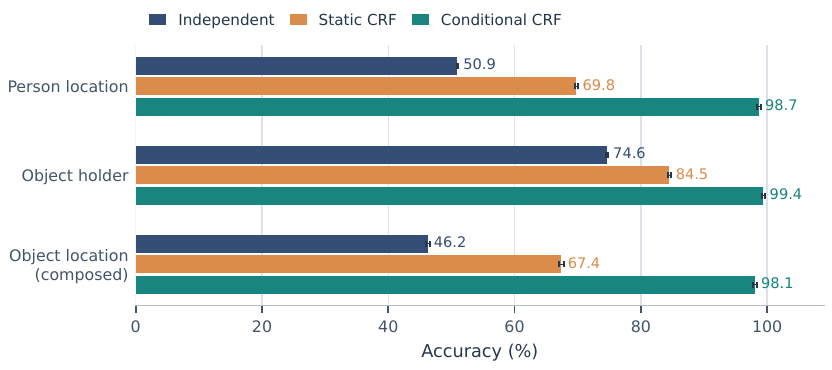}
\caption{Intermediate person-location and object-holder predictions alongside their composed object-location accuracy for MLP decoders with different heads. Bars and error bars show means and standard deviations across three seeds ($n=3$).}\label{fig:intermediate}
\end{figure}
To understand why observation-conditioned transitions are so critical for the per-sentence MLP, Figure~\ref{fig:intermediate} inspects its intermediate predictions. With an independent head, the MLP predicts object holders at $74.6\%$ accuracy but person locations at only $50.9\%$. Conditioning CRF transitions on $h_t$ transforms both chains: person-location accuracy climbs to $98.7\%$ ($99.6\%$ on change steps and $98.5\%$ on persistence steps), and holder accuracy reaches $99.4\%$ ($99.8\%$ on change steps and $99.4\%$ on persistence steps). Because the decoder handles both active updates and silent persistence accurately across intermediate entities, composing them yields $98.1\%$ object-location accuracy.

We next examine how chain inference and composition interact at test time. For the conditional-transition MLP, hard Viterbi composition ($98.1\pm0.3\%$) and soft composition over smoothed marginals ($98.0\pm0.3\%$) are virtually indistinguishable. Crucially, replacing offline smoothing with strictly causal forward filtering and soft composition still achieves $97.3\pm0.5\%$. Because the MLP potentials inspect only the current causal representation $h_t$, this filtered soft variant operates genuinely online on a story prefix without peeking at future events, unlike bidirectional BiLSTM or Transformer encoders, which still rely on future observations even if their CRF head is filtered. For the conditional-CRF BiLSTM, hard Viterbi and soft smoothed composition both reach $99.5\%$ step accuracy.

\subsection{Training data and model families}
\begin{figure}[t]
\centering
\includegraphics[width=\linewidth]{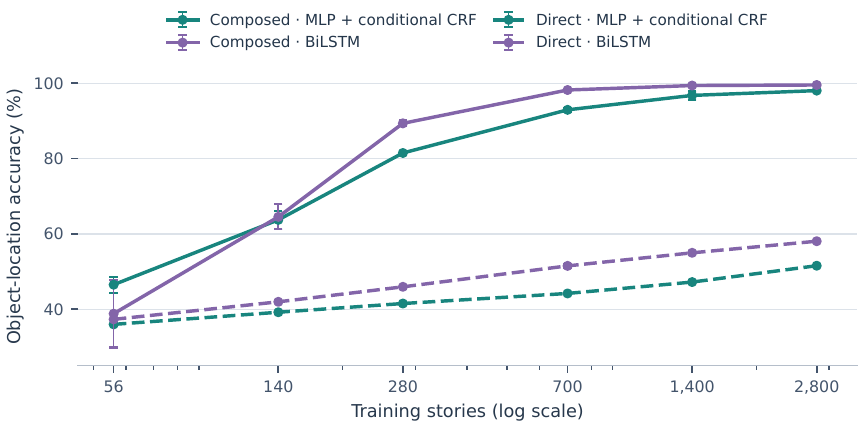}
\caption{Object-location accuracy across training-set sizes. Points and error bars show means and standard deviations across two seeds (subsampled sets) or three seeds (full data). Composed decoders use intermediate supervision and a fixed composition rule; direct decoders train on object locations.}\label{fig:data_efficiency}
\end{figure}
\begin{figure}[t]
\centering
\includegraphics[width=\linewidth]{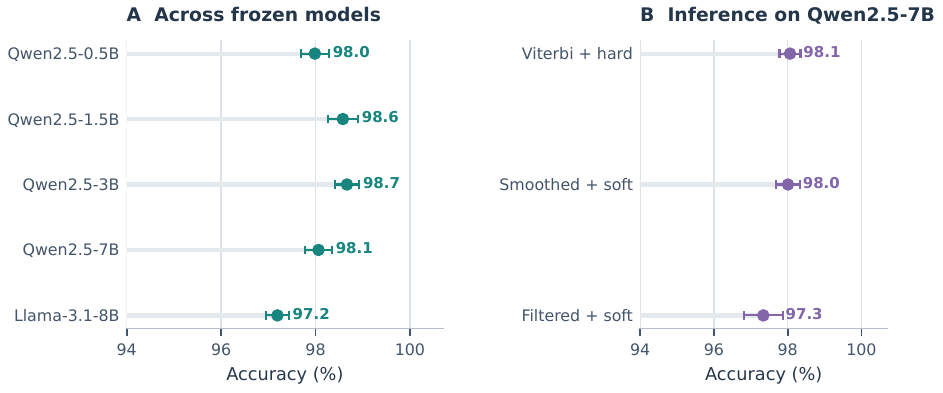}
\caption{Compositional MLP conditional-CRF accuracy across middle-layer representations of five language models (left) and across inference procedures on Qwen2.5-7B (right). Points and error bars show means and standard deviations across three seeds ($n=3$); vertical axes span $94\%$--$100\%$. Only filtered soft composition with per-sentence MLP potentials is strictly prefix-based; the compared inference modes use hard or soft composition, respectively.}\label{fig:models_inference}
\end{figure}
Figure~\ref{fig:data_efficiency} tracks accuracy as training data grows. Even with 280 stories (a tenth of the training set), compositional decoding shows strong sample efficiency: the conditional-CRF MLP reaches $81.5\pm0.6\%$, and the compositional BiLSTM reaches $89.4\pm0.8\%$. At 700 stories, their accuracies climb to $93.0\pm0.5\%$ and $98.2\pm0.1\%$, respectively. Note that at each training step, direct models receive three object-location labels whereas compositional models receive four person-location labels and three holder labels; subsampled runs average two seeds (selecting both the training subset and decoder initialization), while full-data runs use three. On the full training set, whole-story exact match still separates the per-sentence MLP ($82.8\%$) from the conditional-CRF BiLSTM ($93.3\%$), as a single step error anywhere in a story across any of the three objects breaks exact recovery.

Figure~\ref{fig:models_inference} compares feature sources and inference procedures. Across middle-layer features from all five language models examined (Qwen2.5-0.5B, 1.5B, 3B, 7B, and Llama-3.1-8B-Instruct), the compositional MLP conditional-CRF consistently reaches $97.2$--$98.7\%$ object-location accuracy. Moreover, when trained solely on task 2 and evaluated on task 3, where stories average more than three times the length, the compositional BiLSTM conditional-CRF achieves $97.3\pm0.2\%$ accuracy. Detailed breakdowns appear in Appendix~\ref{app:extra}.

\section{Discussion}
Our results point to a clear structural reason why direct object-location decoding lags behind compositional decoding. An object's location follows conditional dynamics: while carried, it inherits its holder's movements; once dropped, it decouples and remains stationary as the holder moves on. Asking a decoder to infer this composite trajectory directly from object labels forces it to learn implicit relational bindings alongside long-range persistence. Factoring the task into person locations and object holders restores two clean, nearly Markovian chains whose interaction is handled exactly by Equation~\eqref{eq:compose}.

Comparing encoder architectures also separates two distinct ways a decoder can model time. In our setup, the CRF head governs compatibility between adjacent states, whereas the potential encoder determines which sentences inform each step's scores. A per-sentence MLP sees only the current hidden state $h_t$ and therefore delegates all cross-sentence memory to the transition layer. This explains why a static transition matrix $A$ falls short ($67.4\%$): global persistence cannot tell when an entity actually moved. Event-conditioned potentials $B_t$ supply that gating mechanism ($98.1\%$). By contrast, a BiLSTM or Transformer aggregates context across the story before scoring, allowing even an independent head to recover accurate trajectories ($99.6\%$), though conditional transitions still tighten whole-story consistency ($93.3\%$ vs.\ $92.3\%$).

From a practical standpoint, separating encoder context from chain inference clarifies the trade-off between online tracking and offline analysis. Pairing per-sentence MLP potentials with forward filtering and soft composition yields a prefix-only decoder ($97.3\%$) that updates state estimates incrementally as each sentence arrives, whereas bidirectional encoders and Viterbi or smoothed inference require full narratives. At the same time, high recovery accuracy under structured decoding does not by itself establish that the frozen language model relies on these states internally, a point we examine in Section~\ref{sec:limitations}.

\section{Limitations}\label{sec:limitations}
High reconstruction accuracy does not prove that a language model internally maintains or computes the decoded world state. In a control using only current-sentence input embeddings rather than contextual hidden states, the compositional MLP--CRF achieves $98.5\pm0.3\%$ accuracy, compared with $98.1\pm0.3\%$ from frozen Qwen2.5-7B layer-14 representations. Because the structured decoder can itself accumulate event cues across steps and execute the relational update rule, strong recovery reflects the joint contribution of the representations, the intermediate supervision, and the decoder's inductive bias. Moreover, direct and compositional decoders differ in supervision, parameter count, and access to a hand-coded composition rule, so their comparison characterizes end-to-end systems rather than a single-factor ablation.

Several properties of the benchmark and evaluation also bound our conclusions. The synthetic bAbI stories use a narrow vocabulary and deterministic event mechanics, and our pooled, deduplicated story split differs from the official benchmark split. Evaluating task-2 decoders on task 3 tests both length generalization and a shift in task distribution. Our main Viterbi and bidirectional results look ahead to future sentences, soft composition relies on a factorized independence approximation, and our experiments cover sentence-final representations from five models whose pretraining exposure to bAbI is unknown. Establishing how models internally compute and use world states will require causal interventions and richer, naturalistic narratives.

\section{Conclusion}
We studied structured decoding of evolving narrative world states from frozen language model representations. By combining intermediate entity supervision, observation-conditioned CRF transitions, and an explicit relational update rule, our compositional decoders achieve accurate and data-efficient trajectory recovery, while forward filtering offers a strictly prefix-based online alternative. More broadly, our experiments highlight the need to account for both decoder inductive bias and supervision when interpreting structured probing results.

\bibliographystyle{plainnat}
\bibliography{references}
\appendix
\section{Implementation details}\label{app:implementation}
The simulator parses movement, pickup, and drop events and saves states after each event. It starts person and object locations as unknown and holders as nobody. Picking up an object can imply a person's location if the object location is already known; moving a person moves objects they hold. Story-level partitions are drawn with seed 0 after within-task deduplication.

The decoder uses dropout 0.1, AdamW learning rate $10^{-3}$, weight decay $10^{-2}$, and gradient-norm clipping at 5. CRF start and transition parameters are initialized from smoothed training-label counts and optimized at learning rate $10^{-2}$ without weight decay. The HMM uses add-one-smoothed training counts and emission scores derived from the independent probe and label priors. Hidden features are stored in float16. The main experiments were run in a four-GPU environment with Python 3.12, PyTorch 2.11, and transformers 5.13. Feature files and model weights are not included in the provided project directory; the extraction and training scripts specify how to regenerate them.

\clearpage
\section{Additional results}\label{app:extra}

\begin{table}[H]
\centering\small
\caption{Compositional MLP + conditional CRF across language models, object-location accuracy (\%, three seeds). Middle layers are used.}\label{tab:models}
\begin{tabular}{lr}
\toprule Model & Accuracy \\
\midrule Qwen2.5-0.5B & $98.0\pm0.3$ \\
Qwen2.5-1.5B & $98.6\pm0.3$ \\
Qwen2.5-3B & $98.7\pm0.3$ \\
Qwen2.5-7B & $98.1\pm0.3$ \\
Llama-3.1-8B-Instruct & $97.2\pm0.2$ \\
\bottomrule
\end{tabular}
\end{table}
\begin{table}[H]
\centering\small
\caption{Training data efficiency, object-location accuracy (\%). Compositional methods use intermediate-state supervision and a fixed rule. Two seeds for subsamples, three for full data.}\label{tab:data}
\begin{tabular}{lrrrr}
\toprule
Method & 280 & 700 & 1,400 & 2,800 \\
\midrule
Composed MLP + cond. CRF & $81.5\pm0.6$ & $93.0\pm0.5$ & $96.8\pm1.1$ & $98.1\pm0.3$\\
Composed BiLSTM & $89.4\pm0.8$ & $98.2\pm0.1$ & $99.4\pm0.0$ & $99.6\pm0.1$\\
\bottomrule
\end{tabular}
\end{table}
\end{document}